\def\year{2021}\relax
\documentclass[letterpaper]{article} 
\usepackage{aaai21}  
\usepackage{times}  
\usepackage{helvet} 
\usepackage{courier}  
\usepackage[hyphens]{url}  
\usepackage{graphicx} 
\usepackage{natbib}  
\usepackage{caption} 
\usepackage{booktabs}
\usepackage{graphicx}
\usepackage{subfigure}
\usepackage{multirow}
\usepackage{paralist}
\usepackage{amsmath}
\usepackage{amssymb}

\title{DirectQE: Direct Pretraining for Machine Translation Quality Estimation}
\author{
    Qu Cui\textsuperscript{\rm 1},
    Shujian Huang\textsuperscript{\rm 1},
    Jiahuan Li\textsuperscript{\rm 1}, 
    Xiang Geng\textsuperscript{\rm 1}, \\
    Zaixiang Zheng\textsuperscript{\rm 1},
    Guoping Huang\textsuperscript{\rm 2}, 
    Jiajun Chen\textsuperscript{\rm 1}\\
}
\affiliations{
    \textsuperscript{\rm 1}National Key Laboratory for Novel Software Technology, Nanjing University, Nanjing, China\\
    \textsuperscript{\rm 2}Tencent AI Lab, Shenzhen, China\\
    \{cuiq, lijh, gx, zhengzx\}@smail.nju.edu.cn, \{huangsj, chenjj\}@nju.edu.cn, donkeyhuang@tencent.com
}

\begin{document}
\maketitle
\begin{abstract}
Machine Translation Quality Estimation (QE) is a task of predicting the quality of machine translations without relying on any reference.
Recently, the predictor-estimator framework trains the predictor as a feature extractor, which leverages the extra parallel corpora without QE labels, achieving promising QE performance. 
However, we argue that there are gaps between the predictor and the estimator in both data quality and training objectives, which preclude QE models from benefiting from a large number of parallel corpora more directly. 
We propose a novel framework called DirectQE that provides a direct pretraining for QE tasks.
In DirectQE, a generator is trained to produce pseudo data that is closer to the real QE data, and a detector is pretrained on these data with novel objectives that are akin to the QE task.
Experiments on widely used benchmarks show that DirectQE outperforms existing methods, without using any pretraining models such as BERT. 
We also give extensive analyses showing how fixing the two gaps contributes to our improvements.
\end{abstract}

\section{Introduction}
Evaluating the results of Machine Translation (MT) usually requires human translations as references, which are expensive to obtain~\citep{papineni2002bleu, meteor}.
Quality Estimation (QE) aims to directly predict the quality of translations without relying on any reference.
An example is shown in Table~\ref{tab:example}.

After those methods relying on hand-crafted features~\citep{Kozlova2016YSDA, Sagemo2016The}, neural QE approaches show promising achievements ~\citep{chen-etal-2017-improving-machine_new,Shah2016SHEF_new,abdelsalam-etal-2016-bilingual_new}.
Because labeled QE data is hard to obtain and often limited in size, it is natural to transfer bilingual knowledge from parallel data of the MT tasks~\citep{kepler2019unbabels}. 
One well-known framework for this knowledge transfer is the \textit{predictor-estimator} framework~\citep{kim-17}.

In the predictor-estimator framework, a \textit{predictor} is trained on parallel data to extract bilingual features.
The training objective is to predict one target token given the source sentence and the rest tokens in the reference~\citep{kepler2019unbabels, QE-BERT}.
Then, an \textit{estimator} will be trained on the amount-limited real QE data to make quality estimations, based on features~\citep{fan2018bilingual} provided by the predictor\footnote{The predictor could also be finetuned on the QE data.}.

\begin{table}[t]
\centering
\footnotesize
\begin{tabular}{l|l}
\toprule
\multirow{2}{*}{Source} & subsequently , each method calls the \\
 & superclass version of itself . \\
\hline
\multirow{3}{*}{Translation} & anschließend wird in jeder Methode \\
 & die \underline{übergeordnete} Superclass-Version\\
 & \underline{von selbst} aufgerufen . \\
\hline
HTER & 0.2727 \\
\bottomrule
\end{tabular}
\caption{
An example of QE. 
Word-level QE assigns labels to each token/word. 
For example, some tokens are tagged as `BAD' (marked underline), and the rest are tagged as `OK'. 
Sentence-level QE labels the whole sentence with a single quality score, representing the effort to correct the translation manually.
}
\label{tab:example}
\end{table}

However, we suggest that the ability of the estimator is limited due to the limited amount of labeled data. On the other hand, there are two major gaps between the predictor and the estimator, which may hinder the transfer of bilingual knowledge.
Firstly, the predictor is trained on the well-translated parallel data for MT. 
In contrast, the estimator aims to work for the QE task, which deals with imperfect translations from MT systems as the input.
The differences in data quality may lead to a degradation of QE performance~\citep{shimodaira2000improving,weiss2016survey}.

Secondly, the predictor is trained with the objective to make token predictions, but the estimator is trained to do QE tasks, which discriminate different translation qualities.
The features used to predict tokens may not be suitable to judge the translation qualities. 
Moreover, during the training of the predictor on parallel data, there is no direct supervision from QE tasks. 
Hence, it is hard for the predictor to learn QE-related features that the estimator can directly exploit.

In this paper, we propose a novel framework called \textit{DirectQE} that provides a direct pretraining for QE tasks.
In DirectQE, a proposed \textit{generator} is first trained on parallel data and then be used to produce a large amount of pseudo QE data.
Our design of the generator makes the generated translations more similar to the real QE data than the parallel data itself and makes it possible to generate the QE labels automatically.
Then a proposed \textit{detector} learns quality estimation by pretraining on the pseudo QE data and finetuning on real QE data.
We design specific pretraining objectives so that the training on the pseudo and real QE data could be well connected.

Unlike the previous estimator, which could only be trained on limited QE data, the detector in our framework enables a direct usage of a large amount of pseudo QE data. 
Besides, instead of relying on features from a QE-irrelevant predictor, our detector performs the quality predictions using features, learned through the direct pretraining, in its own way. 
We conduct extensive experiments to show how fixing these two gaps can contribute to QE tasks. With the same amount of the extra parallel dataset, DirectQE achieves new SOTA results on different QE datasets.

\section{Background}
\subsection{Formalization of QE Task}
Machine Translation Quality Estimation aims to predict the translation quality of an MT system without relying on any reference.
Given a source language sentence $\mathbf{X}$ and a target language translation $\mathbf{T} = \{t_1, t_2, \dots, t_n\}$ with $n$ words, produced by an MT system, the QE system learns to predict two types of quality labels:
\begin{compactitem}
    \item Word-level labels. Label sequence $\mathbf{O} = \{o_1, o_2, \dots, o_n\}$, where $o_j$ is the quality label for the word translation $t_j$, which is usually a binary label meaning `OK' or `BAD'; 
    \item Sentence-level label. A quality score, $q$, for the whole translation $\mathbf{T}$, e.g., Human-targeted Translation Edit Rate (HTER) score~\citep{HTER}. 
\end{compactitem}
Commonly used QE datasets consist of machine translations, as well as the above labels, as tuples $\langle \mathbf{X},\mathbf{T}, \mathbf{O}, q \rangle$.
This kind of label is expensive to get because it requires specific post-editing for the machine translations by experts with bilingual knowledge. Thus human-labeled QE datasets are usually limited in size.

\subsection{Predictor-Estimator Framework}
Since parallel data of the same translation direction is much more available, it is natural to transfer bilingual knowledge from parallel data to QE tasks.
One well-known framework is the predictor-estimator framework~\citep{kim-17}.
It first trains a predictor on parallel data, which acts as a feature extractor. 
The estimator then uses the features from the predictor to make QE predictions. 
Here are two different kinds of predictors in previous research:
\begin{compactitem}
\item NMT-based Predictor.
Training a Neural Machine Translation (NMT) model on parallel data as the predictor~\citep{zhou-cmu, fan2018bilingual, kim-17}.
\item PLM-based Predictor. 
Using a Pretrained Language Model (PLM), e.g., BERT~\citep{devlin2018bert}, as the predictor~\citep{kepler2019unbabels, QE-BERT}. 
\end{compactitem}
The task of both these predictor methods can be seen as a cross-lingual language model, which predicts unseen tokens given the source sentence $\mathbf{X}$ and other tokens in the references $\mathbf{Y}$. 
This task is different from the QE tasks, which aim to predict quality labels.

Once the predictor's pretraining completes, an estimator is trained to incorporate the generation probability of each token in the translation $\mathbf{T}$ as features~\citep{fan2018bilingual} to predict word-level tags and sentence-level scores.

Please note that the estimator can only be trained on the labeled QE data, which is of limited size.
On the other hand, there exist strong gaps between the predictor and estimator, preventing the model from making the best of parallel data.
\begin{figure}[t]
\centering
\includegraphics[width=0.45\textwidth]{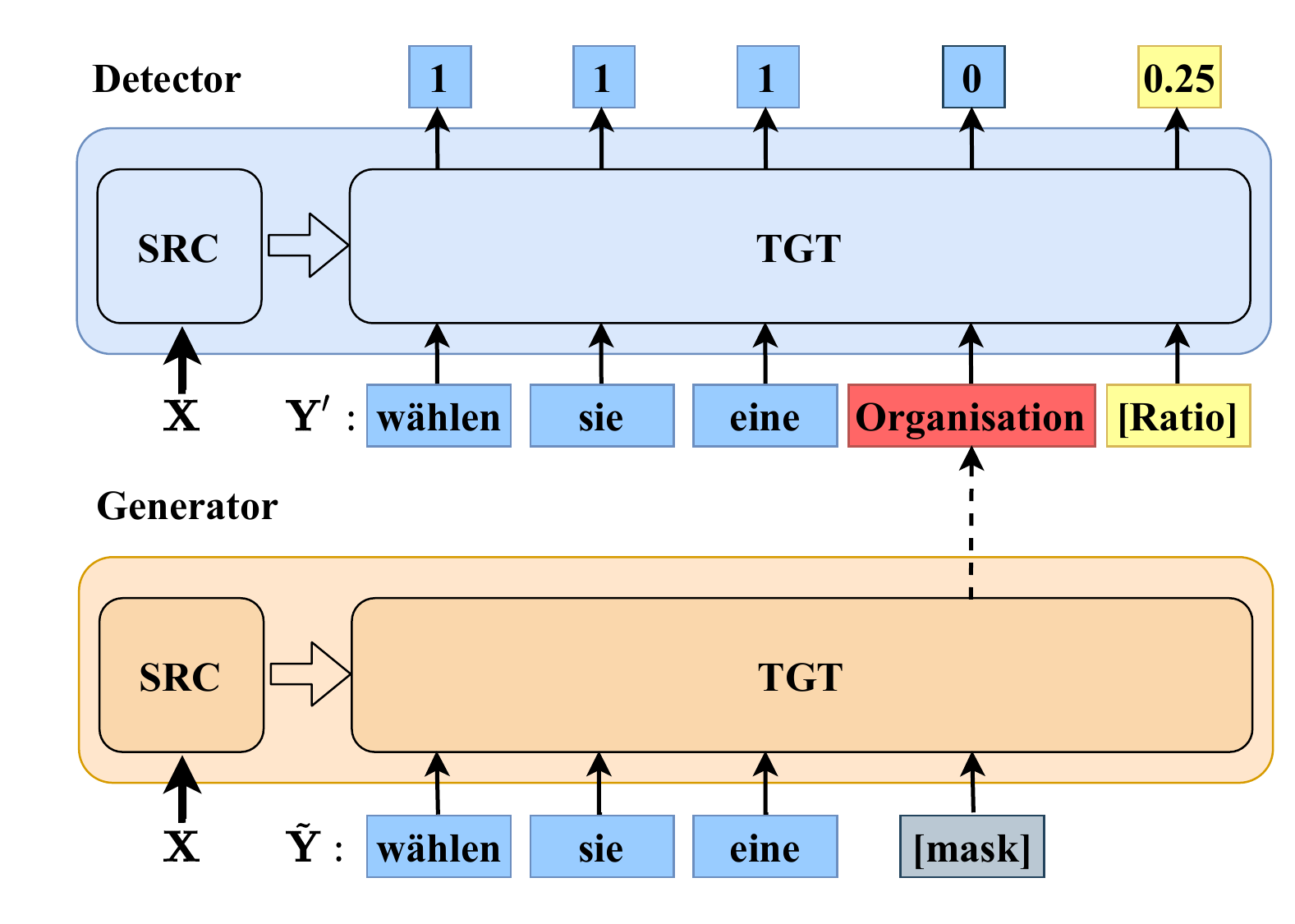}
\caption{
The illustration of our framework. 
The detector is trained on the pseudo QE data from the generator.
In this example, the source sentence is \textit{select a company name} and the parallel translation is \textit{wählen Sie eine Firma}.
}
\label{fig:detector}
\end{figure}

\section{Approach}
To bridge the gaps in the data quality and training objectives between the original predictor and estimator, we propose the DirectQE framework. 
As shown in Figure~\ref{fig:detector}, our framework includes a generator for pseudo QE data generation and a detector pretrained with the same QE objectives on the generated data.
\subsection{Generator}
The generator is trained on the parallel data to generate pseudo QE data.
To achieve this goal, we train the generator as a word-level rewriter.
The word-level rewriter is then used to produce pseudo translations with one-to-one correspondences with the references, from which labels could be automatically generated. 

\subsubsection{Training the Word-level Rewriter on Parallel Data.}
The generator adopts an encoder-decoder architecture like transformer~\citep{NIPS2017_7181}. 
However, it uses a masked language model objective at the target side.
Given a parallel sentence pair $\langle \mathbf{X}$, $\mathbf{Y} \rangle$, 
we generate a masked sentence $\Tilde{\mathbf{Y}}$ by randomly replacing some tokens in the reference with a special tag $\mathtt{[MASK]}$.
The generator is trained to recover the masked tokens given the source tokens and the other tokens at the target side.
Suppose the masked token is $y_j$, then the training objective $\mathcal{J}_{gen}$ is to maximize
\begin{equation}
P(y_j|\mathbf{X}, \tilde{\mathbf{Y}}_{< j}, \mathtt{[MASK]}, \tilde{\mathbf{Y}}_{> j}; \theta).
\end{equation}
Following BERT~\citep{devlin2018bert}, we mask tokens with a 15\% mask ratio.

We do not use a standard NMT model as the generator because it generates translations that may differ significantly from the reference ones. It is usually difficult to get trustworthy quality labels of these translations automatically.

\subsubsection{Generating Pseudo QE Translations.}
We use the trained generator as a word rewriter to produce pseudo QE translations $\mathbf{Y'}=\{y'_1, y'_2, \dots , y'_n\}$.
Notice that the length of $\mathbf{Y'}$ is the same as the reference $\mathbf{Y}$, and the tokens have one-to-one correspondences.

More specifically, we feed parallel data to the generator with some tokens at the target side masked, which is the same as the training process.
For each masked token, the generator will output a probability distribution on the whole vocabulary. 
Unlike the common practice, we do not simply choose the best tokens with the highest generation probability. These tokens' quality in sentences is often too good as a simulation for translation errors.

To generate the pseudo QE translations that share a closer data quality with real QE data than parallel data do, we choose to sample pseudo translations according to the probability distribution.
Here, we introduce the following two different strategies:
\begin{compactitem}
    \item \textbf{Sample}. 
    Similar to~\citet{clark2020electra}, we sample tokens on the whole vocabulary according to the generation probability. 
    To avoid mostly selecting the token with the highest probability, we sample tokens from a softmax with temperature $\tau$~\citep{hinton2015distilling}.
    \item \textbf{Top-k}.
    To improve sampling efficiency, we alternatively randomly select tokens from those with the top $k$ generation probability.
    The value of $k$ affects the overall quality of the translation sentence. 
\end{compactitem}

\subsubsection{Generating Pseudo QE Labels.}
The generator produces a new translation $\mathbf{Y'}$ for each parallel sentence pair.
Then we generate pseudo QE labels on the translation for further training.

Word-level tags $\mathbf{O'}=\{o'_1, o'_2, \dots , o'_n\}$ are generated according to $\mathbf{Y'}$ and $\mathbf{Y}$, which marks a token as `BAD' if it is generated and vice visa:
\begin{align}
    o'_j=\left\{
    \begin{array}{rcl}
    1, & & {\mathtt{if\ }y_j = y'_j\ ,}\\
    0, & & {\mathtt{otherwise.}}
    \end{array} \right.
\end{align}

The sentence-level score $q'$ is generated as the ratio of generated tokens in a translation:
\begin{equation}
q'=1-\frac{\mathtt{sum}(\mathbf{O'})}{\mathtt{len}(\mathbf{O'})}.
\end{equation}

With the above QE labels, we can now get a pseudo tuple $\langle \mathbf{X},\mathbf{Y}^{'}, \mathbf{O}^{'}, q^{'} \rangle$, which shares the same form of real QE data. 
While the real QE dataset is annotated by experts with bilingual knowledge and is quite limited in size, our methods can dynamically generate pseudo QE data at a scale even larger than the parallel data.

\subsection{Detector}
With the pseudo and real QE data, the detector could be pretrained and finetuned, respectively, using the same objectives. That is why we named our method ``direct'' pretraining.
\subsubsection{Pretraining on Pseudo QE Data.}
The detector uses the pseudo QE data produced by the generator for pretraining.
The pretraining task is to jointly predict the tags $O'$ at the word-level and score $q'$ at the sentence-level. 
The pretraining objective of word-level $\mathcal{J}_{w}(\mathbf{X},\mathbf{Y}',o'_j)$ is to maximize
\begin{equation}
\sum_{j=1}^{|\mathbf{O}'|} \text{log}\  P(o'_j | \mathbf{X}, \mathbf{Y}'; \theta),  \label{jw}
\end{equation}
and that of sentence-level $\mathcal{J}_{s}(\mathbf{X},\mathbf{Y}',q')$ is to maximize
\begin{equation}
\text{log} P(q' | \mathbf{X}, \mathbf{Y}'; \theta). \label{js}
\end{equation}

The detector also uses the transformer as the basic framework, but it predicts the quality labels.
It encodes the source sentence with self-attention to obtain hidden representations as the generator does. 
At the target side, the detector predicts word-level tags at each position with the representation from the last encoder layer. 
We also add a special tag $\mathtt{[Ratio]}$ to the translation sequence, and its representation will be used to predict the sentence-level score.

Note that, the \textit{Sample} and \textit{Topk} strategies enable us to generate multiple translations given a parallel sentence pair. In practice, we generate new pseudo QE data for every minibatch. Therefore, the detector could possibly be exposed to diverse translation errors, leading to better QE performance. 

It is important to control the pretraining process for the detector.
We randomly cut 2,000 sentence pairs out of the parallel data and use the generator to produce pseudo QE data with the \textit{Sample} strategy.
This dataset is used as the development set to monitor the pretraining process, and the performance on this dataset will be used for model selection.

\begin{table*}[!t]
\centering
\footnotesize
\begin{tabular}{c|l|ccc|ccc|c|c}
\toprule
\multirow{2}{*}{Dataset} & 
\multirow{2}{*}{Method} & 
\multicolumn{3}{c|}{Sent-level Dev} & 
\multicolumn{3}{c|}{Sent-level Test} & 
Word-level Dev & 
Word-level Test 
\\
&  & Pearson$\uparrow$ & MAE$\downarrow$ & RMSE$\downarrow$ & Pearson$\uparrow$ & MAE$\downarrow$ & RMSE$\downarrow$ & F1-MULT$\uparrow$ & F1-MULT$\uparrow$ \\ 
\hline
\hline
\multirow{4}{*}{WMT19} &
  NMT-based  & 53.67 & 10.97 & 16.06 & 50.65 & 11.92 & 16.77 & 15.74 & 17.08 \\
& PLM-based* & 55.10 & 10.39 & 15.94 & 52.08 & 11.39 & 16.66 & 37.03 & 37.04 \\
\cline{2-10}
& DirectQE-Sample  &  \textbf{57.99} & 10.16 & 15.75 & 54.41 & 11.36 & 16.56 & 38.27 & 36.60 \\
& DirectQE-Top-k   &  57.19 & \textbf{10.14} & \textbf{15.72} & \textbf{55.08} & \textbf{11.25} & \textbf{16.33} & \textbf{40.63} & \textbf{39.71} \\
\hline
\hline
\multirow{4}{*}{WMT17} &
  NMT-based  & 68.62 & 10.74 & 15.44 & 67.44 & 10.65 & 14.48 & 44.43 & -  \\
& PLM-based* & 72.48 & \textbf{9.85} & \textbf{14.07} & 72.01 & 9.89 & 13.30 & 57.58 & -\\
\cline{2-10}
& DirectQE-Sample  & 72.78 & 9.95  & 14.97 & 72.45 & 9.71  & 13.52 &  \textbf{58.18} & - \\
& DirectQE-Top-k   & \textbf{73.38} & 9.92  & 14.72 & \textbf{73.56} & \textbf{9.51}  & \textbf{13.15} &57.99 & - \\
\bottomrule
\end{tabular}
\caption{
Main results of NMT-based QE, PLM-based QE, and our DirectQE on two EN-DE QE datasets.
\textit{Sample} means we select tokens on the whole vocabulary according to the generation probability. 
\textit{Top-k} means we randomly select one of the tokens with the top $k$ generation probability.
`-' indicates missing test results on WMT17, because the word-level golden tags are not released yet.
`*' indicates systems that use additional resources such as BERT.
}
\label{tab:main results}
\end{table*}

\subsubsection{Finetuning on Real QE Data.}
After pretraining, the detector will be finetuned on real QE tasks directly.
The objectives of finetuning are $\mathcal{J}_{w}(\mathbf{X},\mathbf{T},o_j)$ and $\mathcal{J}_{s}(\mathbf{X},\mathbf{T},q)$ using real data and labels from QE datasets instead of pseudo ones.

When pretraining the detector, we do not finetune the generator together since it is difficult to back-propagate through sampling from the generator~\citep{clark2020electra}.

Unlike existing methods, with the help of the generator, we can pretrain the detector with QE objectives on a large amount of pseudo QE data, instead of training a language model task on parallel data.
Furthermore, the detector can be directly used to do real QE tasks instead of introducing an extra estimator.

\section{Experiments}
\subsection{Experimental Settings}
\subsubsection{Dataset.}
We carry out experiments on the WMT19 and WMT17 QE tasks for English-to-German (EN-DE) direction. 
The translation of the WMT19 task is generated by an NMT system, while in WMT17 a statistical MT system is used.
The EN-DE parallel dataset is from the WMT19 Shared Task and contains 3.4M sentence pairs, which are far larger than the QE dataset (about 13k).
These datasets are all officially released for the WMT QE Shared Task.

\subsubsection{Models.}
In our experiments, we reproduce the two best previous models as our baselines and compare them with the proposed DirectQE.
\begin{compactitem}
    \item NMT-based QE. Using an NMT model as the predictor, and a Bi-LSTM~\citep{GravesFramewise} as the estimator, following the architecture of QE Brain~\citep{fan2018bilingual}.
    \item PLM-based QE. Using BERT~\citep{devlin2018bert} as the predictor, and a Bi-LSTM~\citep{GravesFramewise} as the estimator.
    The code and pretrained parameters are from the huggingface~\citep{Wolf2019HuggingFacesTS}. Please note that BERT is pretrained with a large number of external resources.
    \item DirectQE. We implement the models described in the previous section with the \textit{Sample} and \textit{Top-k} strategies.
\end{compactitem}
When training the estimator, the model is jointly trained to predict the word-level and sentence-level QE labels.

\subsubsection{Implementation Details.}
For NMT-based QE, The predictor of NMT-based QE consists of an encoder and a decoder, each of them is a 6-layer Transformer with hidden states dimension 512. The estimator is a single layer Bi-LSTM with hidden states dimension 512.
The total number of parameters is about 113M.
    
For PLM-based QE, the encoder is the same as the released model, which is a 12-layer Transformer with hidden states dimension 768. The estimator is a single-layer Bi-LSTM with hidden states dimension 768.
The total number of parameters is about 177M.

For DirectQE, the detector is based on the transformer~\citep{NIPS2017_7181}, with one encoder and one decoder, also the same size as the NMT-based QE. 
The generator is based on a transformer of 6 layers but with hidden states dimension 256 for each layer~\citep{clark2020electra}. 
The total number of parameters is about 90M.

Please note that our DirectQE is the smallest in size among all three architectures.

We use BPE~\citep{sennrich2015neural} in our experiments and set BPE steps to 30,000. 
For the word-level task, if subtokens of a word have conflicting prediction results, we simply consider its word-level label as `BAD'.

\subsubsection{Metrics.}
The evaluation of our experiments follows the WMT QE shared task.
For the word-level task, the metric is the product of the F1-score for the `OK' and `BAD' tokens (F1-MULT).
For the sentence-level task, the main metric is Pearson's Correlation Coefficient. 
We also use mean absolute error (MAE), and root-mean-square error (RMSE).

\subsection{Main Results}
We use the same parallel data and QE data to train all the systems except for the PLM-based QE, where a pretrained BERT is employed, as described in \citep{QE-BERT}.  The results are reported in Table~\ref{tab:main results}.

The PLM-based QE performs better than the NMT-based QE, possibly because it contains extra knowledge from the monolingual corpora where BERT is trained.
DirectQE outperforms the others on the test datasets at both the word-level and sentence-level.
Please note that DirectQE does not use any extra monolingual data and has the least model parameters among the three models. 
Both the two sampling strategies of DirectQE achieve remarkable performance.
We will discuss more on this in Analysis.

\begin{figure}
    \centering
    \includegraphics[width=0.42\textwidth]{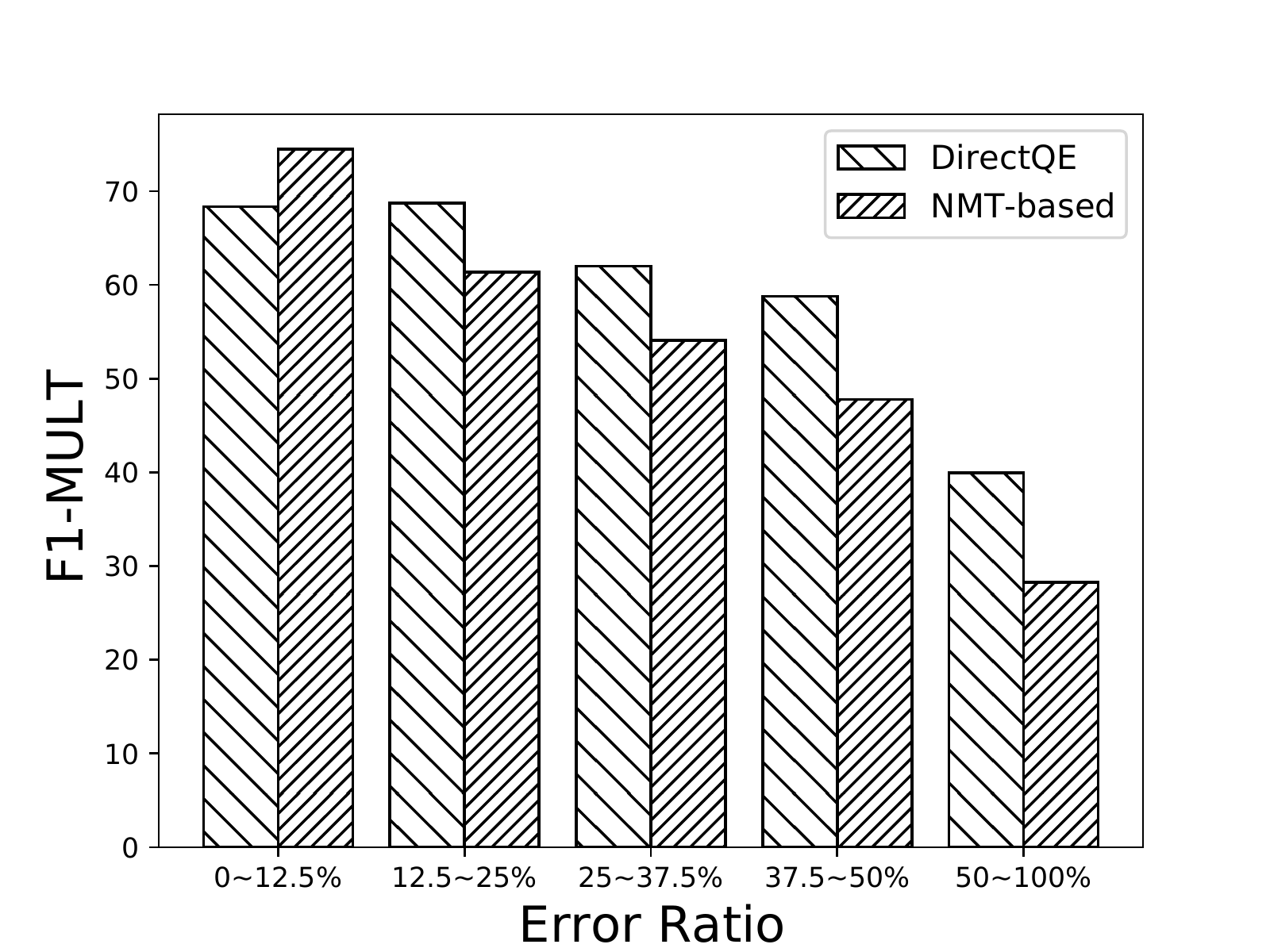}
    \caption{
    F1-MULT of our method against the NMT-based method on the data with different error ratios.
    }
    \label{fig:WMT17 word-level}
\end{figure}

To figure out the detailed improvement of our model, we divide the real QE data according to their qualities, i.e. the ratio of `BAD' tokens in translations.
Then we compare the word-level performance of our method and the NMT-based method on each part of the data (Figure~\ref{fig:WMT17 word-level}). 
When the error ratio is larger than 12.5\%, our method achieves better performance than the NMT-based method. 
The relative improvement is much larger when the error ratio becomes larger, showing that the advantage of our method is stronger when the translation quality is lower. 
This improvement could be attributed to the introduction of pseudo QE data with our generator, which is closer to the real translations.

\subsection{Results of Ensemble Models}
Instead of the single system results reported in the previous sub-section, many existing works of QE tasks only report their results of ensemble models. 
These results are hard to compare because different research may use different ensemble methods or use different data.
To get a relatively clear comparison, we only employ a very simple ensemble method: we perform a weighted summation on all the test results from our models with different settings and hyper-parameters, where the weights are learned from the development set~\citep{kepler2019unbabels}.

As shown in Table~\ref{tab:ensemble results}, our system can achieve better ensemble performance on both two datasets.
We want to point out that the existing best ensemble system uses seven different methods~\citep{kepler2019unbabels}, while our system only ensembles the three methods mentioned in our work. 
Since the DirectQE is conceptually distinct from the other two methods, and it is well known that diversity of models matters for ensemble~\citep{krogh1995neural}, we expect our model could be beneficial for promoting the development of better QE systems.

\subsection{Results on Other Language-pairs}
We also conduct experiments on the WMT19 EN-RU and WMT20 EN-ZH QE tasks.
For the EN-RU language pair, we use the official parallel dataset (12M) from the WMT19 QE task.
For the EN-ZH language pair, we use the parallel dataset (7.5M) from the WMT18 translation task.
As shown in Table~\ref{tab:en-ru and en-zh qe}, DirectQE can also achieve better performance on both datasets.

\section{Analysis}
\label{section5}
We present analyses of our methods from two perspectives: data quality and training objectives.
\begin{figure*}[htbp]
\centering
\subfigure[]{
\begin{minipage}[t]{0.5\textwidth}
\centering
\includegraphics[width=0.84\textwidth]{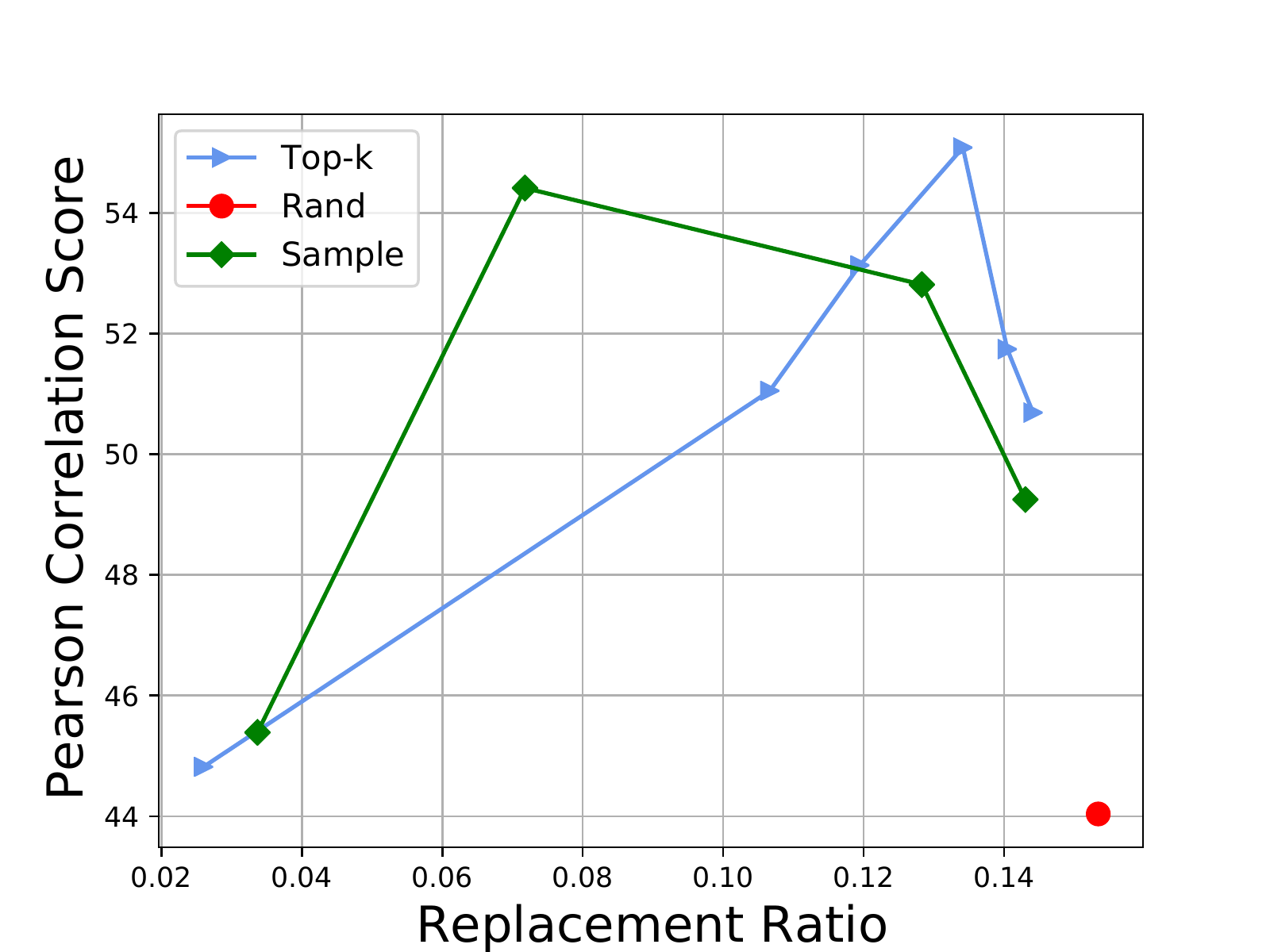}
\label{fig:topk}
\end{minipage}%
}%
\subfigure[]{
\begin{minipage}[t]{0.5\textwidth}
\centering
\includegraphics[width=0.84\textwidth]{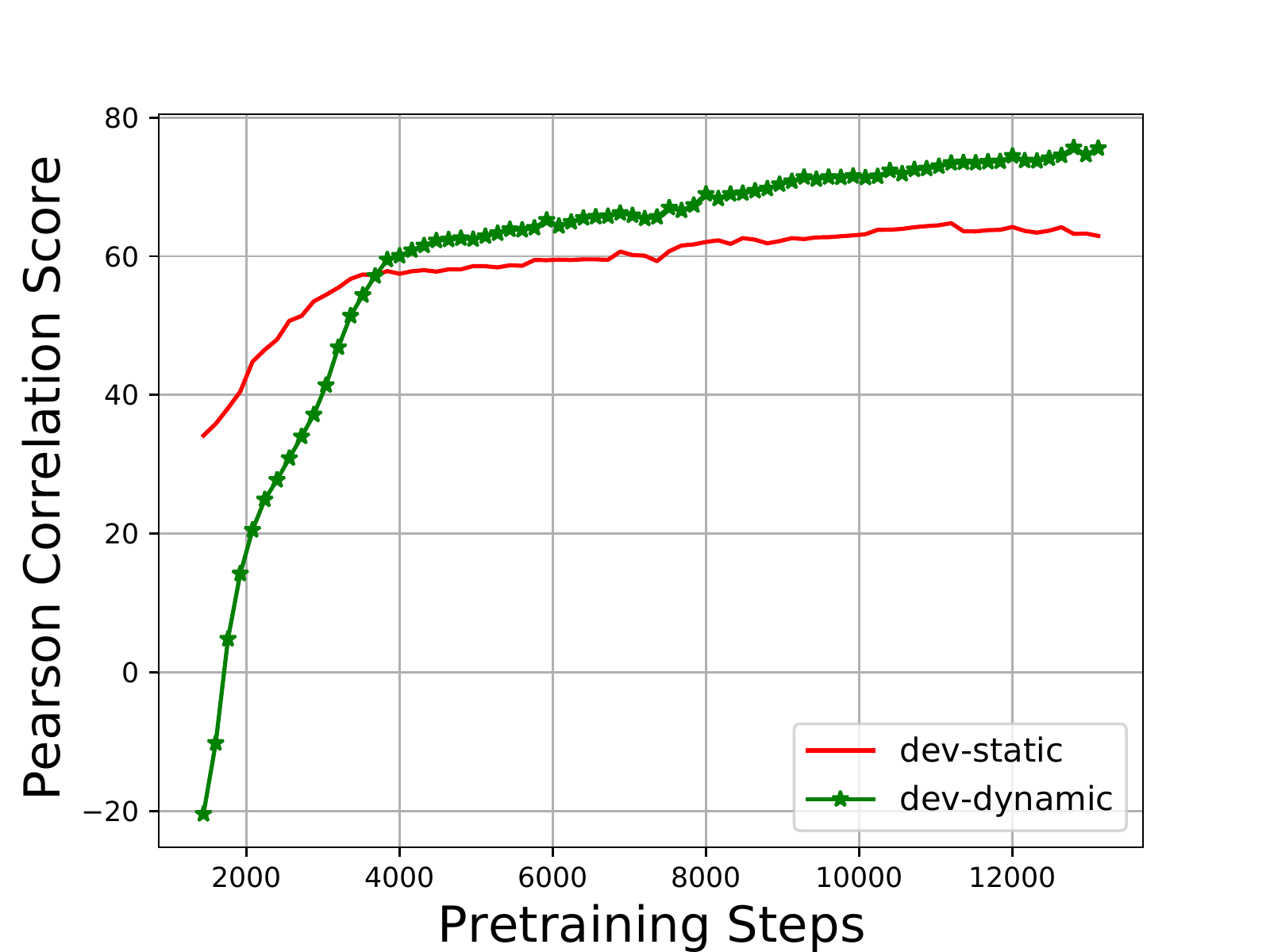}
\label{fig:static_dynamic}
\end{minipage}
}%
\caption{
(a) QE performances according to different replacement ratios. 
The Top-k curve is models with k equals to 1, 3, 5, 10, 20, 40. 
The Sample curve is models with temperature equals to 1, 2, 4, 8. 
Rand is the model with random replacement.
(b) The results on the pseudo QE development set according to the time steps.
}
\end{figure*}

\begin{table}[tbp]
\centering
\footnotesize
\begin{tabular}{c|l|cc}
\toprule
Dataset & System & Sent-level Dev & Sent-level Test \\
\hline
\multirow{4}{*}{WMT19}
& QE-BERT  & 54.50 & 52.60\\
 & SOURCE  & -     & 54.74 \\
 & UNBABEL & 59.68 & 57.18 \\
\cline{2-4}
 & DirectQE & \textbf{60.95} & \textbf{57.25} \\
\hline
\hline
\multirow{4}{*}{WMT17}
& UNBABEL  & 64.33 & 64.10\\
& POSTECH  &     - & 69.54 \\
& QE-Brain &     - & 71.59 \\
\cline{2-4}
& DirectQE & \textbf{77.63} & \textbf{76.29} \\
\bottomrule
\end{tabular}
\caption{
Results of ensemble models for the sentence-level QE.
We present results reported in their original papers~\citep{QE-BERT,zhou-cmu,kepler2019unbabels,Martins2017Unbabel,kim-17,fan2018bilingual}.
`-' marks missing results, which are not reported. For the sake of brevity, we only report the Pearson Correlation Score for the sentence-level evaluation in this Table, Table~\ref{tab:en-ru and en-zh qe}, Table~\ref{tab:data1},  Table~\ref{tab:data2} and Figure~\ref{fig:topk},\ref{fig:static_dynamic}.
}
\label{tab:ensemble results}
\end{table}

\begin{table}[tbp]
    \centering
    \footnotesize
    \begin{tabular}{c|c|cc|cc}
    \toprule
    \multirow{2}{*}{Dataset} & 
    \multirow{2}{*}{Method} & 
    \multicolumn{2}{c|}{Sent-level} & 
    \multicolumn{2}{c}{Word-level}\\
    & & Dev & Test & Dev & Test \\
    \hline
   \multirow{3}{*}{EN-RU}
   & NMT-based & 43.85 & 50.16 & 26.85 & 27.03 \\
   & PLM-based & 43.94 & 52.94 & 23.35 & 26.44 \\
   \cline{2-6}
   & DirectQE & \textbf{48.80} & \textbf{53.34} & \textbf{30.84} & \textbf{35.75}\\
    \hline
   \multirow{3}{*}{EN-ZH}  
   & NMT-based & 58.05 & 56.60 & 44.07 & 43.26 \\
   & PLM-based & 61.98 & 59.54 & 47.82 & 47.91 \\
   \cline{2-6}
   & DirectQE & \textbf{62.48} & \textbf{60.83} & \textbf{52.46} & \textbf{53.06}\\
    \bottomrule
    \end{tabular}
    \caption{Results on the EN-RU and EN-ZH dataset.}
    \label{tab:en-ru and en-zh qe}
\end{table}

\subsection{Data Quality}
Compared with parallel data, our generator produces translations with some errors, which is more similar to the quality estimation data. We first conduct experiments to show the effects of these data quality differences. 

\subsubsection{Pseudo QE Data is Better than Parallel Data.}
Instead of using the detector, we keep the predictor-estimator architecture and train the NMT-based QE system.
The performance is compared for the NMT-based systems trained with pseudo translations produced by our generator, and trained with the original parallel data.
Table~\ref{tab:data1} shows the results. 
We can see that the performance is improved in most of the comparisons, demonstrating the benefits of generated pseudo QE data.

\subsubsection{Quality of Pseudo Translations is Important.}
\label{sec:hyper-para}
During the training of the generator, strategies and hyper-parameter choices may affect the quality of the generated pseudo translation. We present experiments to study these effects. 
We notice that the different choices of strategies and hyper-parameters mainly affect the ratio of replacement during the pseudo translation generation process. 
In the Sample strategy, the higher the temperature is, the more likely the current word is replaced. In the Top-k strategy,  a hyper-parameter $k$ controls the replacement choices in generated pseudo translations, where higher $k$ values indicate a higher replacement ratio.

Thus, we plot the performance of sentence-level quality estimation with the replacement ratio in Figure~\ref{fig:topk}. 
As the replacement rate increases, the quality of pseudo-translations becomes worse and gradually approaches the real translation, so the QE performance first improves.
However, when the replacement ratio is too high, the quality pseudo-translations may be worse than real translations which is clear harm to the QE results.
In both strategies, there is a balance between pseudo data quality and QE results.
The best performance is achieved at temperature 2 for the Sample strategy and top-10 for the Top-k strategy.

It could be easily seen that with a random replacement, the QE performance is quite weak (bottom-right corner) because the quality of the replacement cannot be ensured.

\begin{table}[!tbp]
    \centering
    \footnotesize
    \begin{tabular}{c|c|cc|cc}
    \toprule
    \multirow{2}{*}{Dataset} & 
    \multirow{2}{*}{Training} & 
    \multicolumn{2}{c|}{Sent-level} & 
    \multicolumn{2}{c}{Word-level}\\
    & & Dev & Test & Dev & Test \\
    \hline
   \multirow{2}{*}{WMT19} 
   & parallel  & 53.67 & 50.65 & 15.74 & \textbf{17.08} \\
   & pseudo &  \textbf{54.11} & \textbf{51.50} & \textbf{18.85} & 16.69 \\
    \hline
   \multirow{2}{*}{WMT17}  
   & parallel  & 68.62 & 67.44 & 44.43 & -\\
   & pseudo & \textbf{69.45} & \textbf{68.19} & \textbf{45.99} & -\\
    \bottomrule
    \end{tabular}
    \caption{
    Results of the the NMT-based QE system trained on the generated sentence pairs against those on the parallel sentence pairs.}
    \label{tab:data1}
\end{table}

\subsubsection{Diversity of Data Matters.}
\label{unlimited}
During the pretraining of detector, we could re-sample from the generator distribution every epoch and generate the pseudo QE data dynamically. This enables us to train the detector with larger and more diverse data. 
To understand how this could affect the final performance, we use the generator to produce a fixed set of pseudo QE data, which is of the same size as the parallel data. We then compare the procedures of pretraining the detector on this fixed (static) set v.s. on the dynamically generated QE data.

As shown in Figure~\ref{fig:static_dynamic}, the detector pretrained on the static set converges more quickly, but that pretrained on the dynamic data has better performance on the development set at last.
Considering the gap of performance when the pretraining ends, it is obvious that the dynamic generated pseudo data is necessary for the detector.

\subsection{Training Objective}
Another gap in the predictor-estimator framework is the training objectives. We conduct the following experiments about DirectQE and the NMT-based system on the same pseudo QE data to investigate the influence of this gap.

\begin{figure}
    \centering
    \includegraphics[width=0.42\textwidth]{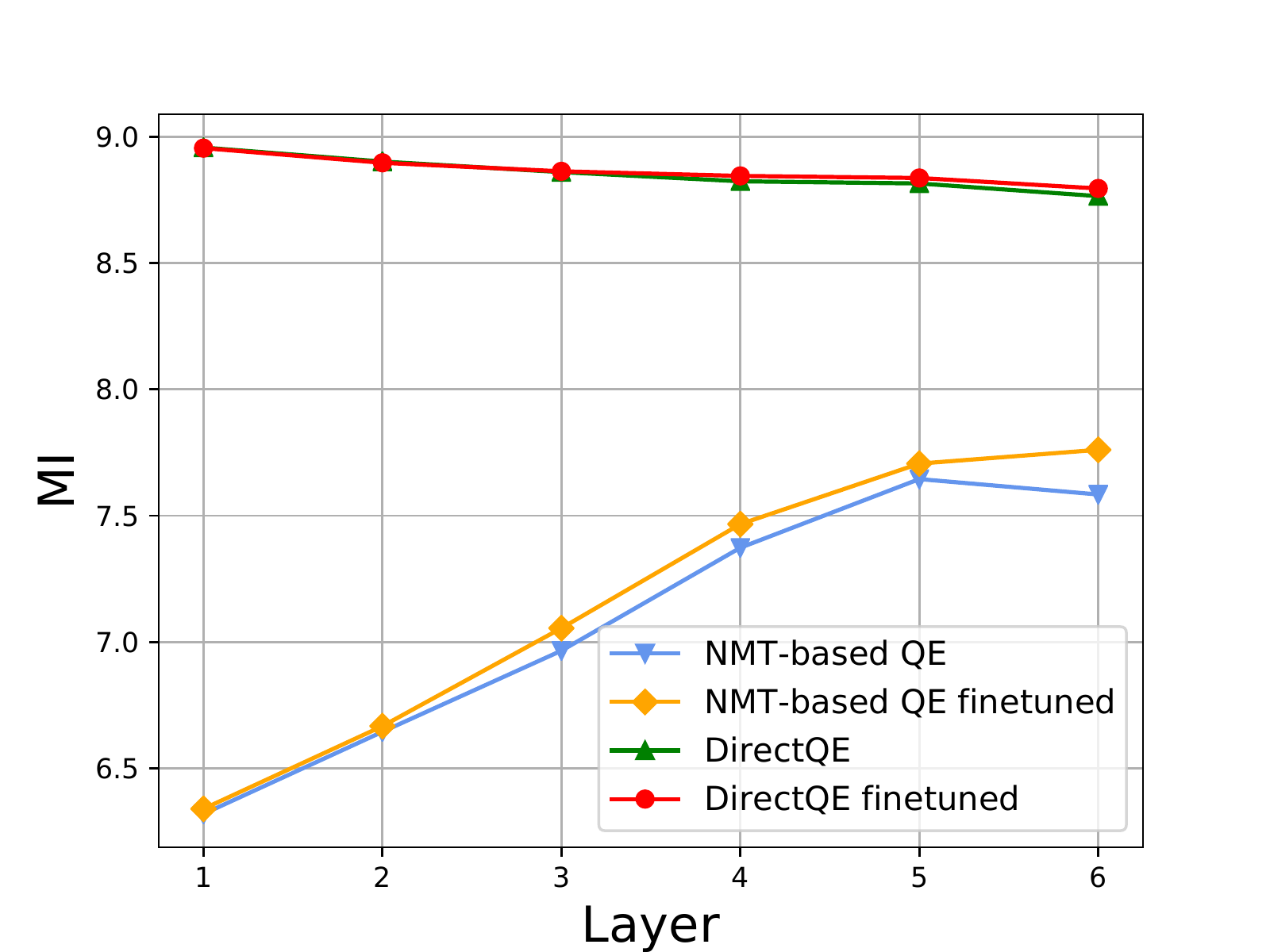}
    \caption{
    The mutual information between the representations and current tokens at the target side.
    }
    \label{fig:mi}
\end{figure}

\subsubsection{The Detector Representation Contains More Information about the Current Tokens.}
Before making quality decisions (especially for word-level decisions), it is vital to gather enough information about the current tokens.
We design experiments to measure the correlation between final representations and current tokens. 
For DirectQE, the final representations refer to the representations in each layer of the detector; for NLP-based systems, the final representations refer to the representations in each layer of the decoder part of the predictor. Our measurement is based on mutual information (MI)~\citep{estimate_mi, IBMI}. 
The bigger the MI value, the more information about current tokens in the final representation.

To estimate the MI, we follow the way described in~\citet{estimate_mi}, which uses the clustering method to discretize representations.
We gather representations of the 100 most frequent tokens at each layer of both models and cluster these representations into 1000 cluster centers using $k$-means.
Mutual information is represented by the co-occurrence frequency of the tokens and centers.

As shown in Figure~\ref{fig:mi},  
the NMT-based system does not contain much information on current tokens at lower layers, and then the MI becomes higher in higher layers. On the other hand, DirectQE has much higher MI between current tokens and representations at each layer than the NMT-based system.
This difference clearly shows the influence of different training objectives. And our proposed method learns a better representation of current tokens and is more capable of making quality predictions. 

\subsubsection{Pretraining with the QE-like Objectives Learns More Suitable Representations for QE Tasks.}
We hypothesize that representations after finetuning are better for the downstream tasks~\citep{howard2018universal}.
Therefore, in the process of finetuning the downstream task, the smaller the representation changes, the more suitable the initial representation is for the task. 

We measure the similarity representations before and after finetuning for DirectQE and NMT-based QE, respectively, where a higher similarity may indicate better representations.
We get all representations of QE pairs from the source side and target side and calculate their similarity by centered kernel alignment (CKA)~\citep{kornblith2019similarity} and canonical correlation analysis (CCA)~\citep{raghu2017svcca}. 

Table~\ref{tab:rep} shows that the representations of DirectQE have higher similarity before and after finetuning, on three different metrics at both the source and target side.
We can also find a similar phenomenon in Figure~\ref{fig:mi}, that after finetuning on real QE data, our method does not change a lot while the NMT-based system has a bigger difference.
It demonstrates that our detector can learn better representations for QE tasks at the pretraining stage, leading to better final QE performance.

\begin{table}[t]
\centering
\footnotesize
\begin{tabular}{c|c|c|c}
\toprule
Metrics & Methods & Source$\uparrow$ & Target$\uparrow$ \\
\hline
\multirow{2}{*}{Linear CKA} & NMT-based QE & 0.8106 & 0.2415 \\
& DirectQE & \textbf{0.9815} & \textbf{0.8693} \\
\hline
\multirow{2}{*}{RBF CKA} & NMT-based QE & 0.8480 & 0.4838 \\
& DirectQE & \textbf{0.9478} & \textbf{0.7503} \\
\hline
\multirow{2}{*}{SVCCA} & NMT-based QE & 0.9469 & 0.8290 \\
& DirectQE & \textbf{0.9711} & \textbf{0.8626} \\
\bottomrule
\end{tabular}
\caption{Similarity of representations before and after finetuning.}
\label{tab:rep}
\end{table}

\subsubsection{Sentence-level Pretraining Does Help.}
We generate both sentence-level and word-level QE labels for our pseudo QE data. We provide an experiment to evaluate the importance of sentence-level pretraining.
As shown in Table~\ref{tab:data2}, the performance is greatly affected without sentence-level pretraining, which means the training objectives of pretraining and downstream tasks should be strictly consistent.

\begin{table}[t]
\footnotesize
\centering
\begin{tabular}{c|cc}
\toprule
Dataset & Sent-level & Word-level\\
\hline
WMT19 & 51.03\ ($-$4.05) & 37.10\ ($-$2.61) \\
WMT17 & 72.55\ ($-$1.01) & - \\
\bottomrule
\end{tabular}
\caption{Results of DirectQE-Top-k on two test sets without the sentence-level pretraining objective.}
\label{tab:data2}
\end{table}

\section{Related Work}
The combination of a generator and a detector looks like Generative Adversarial Network (GAN)~\citep{caccia2018language}. 
However, we do not train these two parts together as GAN does.
The usage of the generator is only to produce pseudo QE data for pretraining the detector.

Our method has a similar model architecture with the Electra~\cite{clark2020electra}.
We both use a generator to produce pseudo sentences, and use a detector to find out the generated tokens in these sentences.
However, we do not share the same motivation nor task.
Their work improves on predictive pretraining for the masked language model, which lacks efficiency for general scenarios of natural language understanding, while a discrimination task may be more efficient.
Our motivation is to bridge the gap between pretraining on parallel data and finetuning on QE data and build a direct pretraining method for QE tasks.
Meanwhile, their model is monolingual and can not be used to solve QE tasks.

We also share the same idea of using pseudo data with some automatic post-editing (APE) work, which tries to use different MT systems to produce pseudo APE data~\citep{negri2018escape, junczysdowmunt2016loglinear}.
It is similar to what we do that we train the generator on parallel data to produce pseudo data.
The difference is that they directly use MT systems to generate machine translations, while we modify ground-truth translations into corrupted but natural sentences, where the modified one-to-one corresponded tokens can be inferred to provide QE-oriented learning signals.
That is, we require data that can be annotated automatically and trustworthy.

\section{Conclusion}
We propose a novel architecture called DirectQE that provides a direct pretraining for QE tasks.
In our method, a generator is first trained on parallel data and will be used to produce the pseudo QE data.
Then a detector will be directly pretrained with quality estimation on these pseudo data and then finetuned on real QE data.
Compare with previous methods, our method can bridge the gaps in the data quality and training objectives between the pretraining and finetuning, which enables our model to learn more suitable knowledge for QE tasks from parallel data.
Extensive experiments show the effects of each part of our method.

In future work, it is interesting to investigate automatically obtaining QE labels for naturally generated translations.
Meanwhile, currently, we can only generate errors with substitution, and it is also interesting to simulate insertion and deletion operations and investigate their influences on translation quality.

\section*{Acknowledgements}
We would like to thank the anonymous reviewers for their insightful comments. 
Shujian Huang is the corresponding author. 
This work is supported by National Key R\&D Program of China (No. 2019QY1806), National Science Foundation of China (No. 61772261). 
This work is also supported by research funding from Huawei Corporation (No. YBN2020115023).

\end{document}